\documentclass[11pt]{style/prism-princeton}

\usepackage{capt-of}
\usepackage{subcaption}
\usepackage{wrapfig}

\definecolor{ctrlf_purple}{HTML}{756BD9}
\colorlet{princeton_orange}{ctrlf_purple}
\colorlet{bgcolor}{ctrlf_purple!8}
\hypersetup{linkcolor=ctrlf_purple,citecolor=ctrlf_purple,urlcolor=ctrlf_purple}

\newif\ifdraftnotes
\draftnotestrue 

\ifdraftnotes
  \newcommand{\ycnote}[1]{\textcolor{magenta}{\footnotesize [YC: #1]}}
  \newcommand{\wnote}[1]{}
  \newcommand{\sgoal}[1]{\par\noindent\textcolor{teal}{\footnotesize [GOAL: #1]}\par}
\else
  \newcommand{\ycnote}[1]{}
  \newcommand{\wnote}[1]{}
  \newcommand{\sgoal}[1]{}
\fi

\author[*]{Mengjie Zhao}
\author[*]{Yuxin Chen}
\author{Yiyang Shao}
\author{Chuye Hong}
\author{Jiakang Chen}
\author{Koushil Sreenath}
\author{Masayoshi Tomizuka}

\affiliation{University of California, Berkeley}
\contribution[*]{Equal contribution.}

\begin{document}

\title{\raisebox{-0.35em}{\includegraphics[height=1.4em]{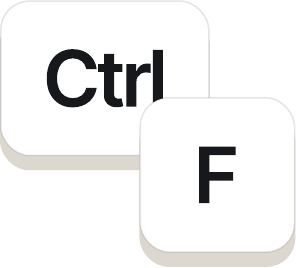}}\hspace{0.3em}\parbox[t]{0.9\linewidth}{\setstretch{1.1}Ctrl-F: Force as an Interface for\\Humanoid Loco-Manipulation}}

\abstract{
Humanoid motion tracking has scaled to general policies that reproduce
diverse human motion, but the resulting behavior is in free-space: the tracker
knows where the body should be, not how it should interact with the
environment. Other work makes the object part of the objective, yet abstracts
that interaction to contact or geometry, which leaves the demanding cases
under-explored. We argue that force is a natural command channel between motion imitation
and object interaction, one that specifies not just where to make contact but
how hard.
We present \textbf{Ctrl-F}, a hierarchical motion--force framework that
augments motion tracking with an explicit force command, a 3-D force at each
designated contact body such as a hand or a foot. The hierarchical framework comprises:
\textbf{(1)} A unified motion--force controller tracks the reference motion
while applying the commanded force to the world.
\textbf{(2)} A high-level force planner that issues a force command online based on
object state and proprioception observations.
In simulation and on hardware, Ctrl-F completes loco-manipulation that demands
large, sustained interaction force and adjusts its command online as the load
changes, both out of reach of a force-blind tracker. 
On hardware, Ctrl-F demonstrates hand-based dragging, pushing, and lifting, as well as foot-based kicking, successfully manipulating boxes weighing up to $20$\,kg in pushing and kicking tasks where the force-blind baseline fails, while adapting online to additional loads introduced mid-task.
}


\website{https://forcectrl.github.io/}{forcectrl.github.io}

\providecommand{\codelist}{}

\makeatletter
\let\ctrlf@maketitle\maketitle
\renewcommand{\maketitle}{%
  \begingroup
  \let\ctrlf@includegraphics\includegraphics
  \renewcommand{\includegraphics}[2][]{%
    \ifnum\pdfstrcmp{##2}{style/logo/princeton_logo.png}=0
    \else\ctrlf@includegraphics[##1]{##2}\fi}%
  \ctrlf@maketitle
  \endgroup}
\makeatother

\maketitle

\begin{figure}[!tp]
\centering
\includegraphics[width=0.75\linewidth]{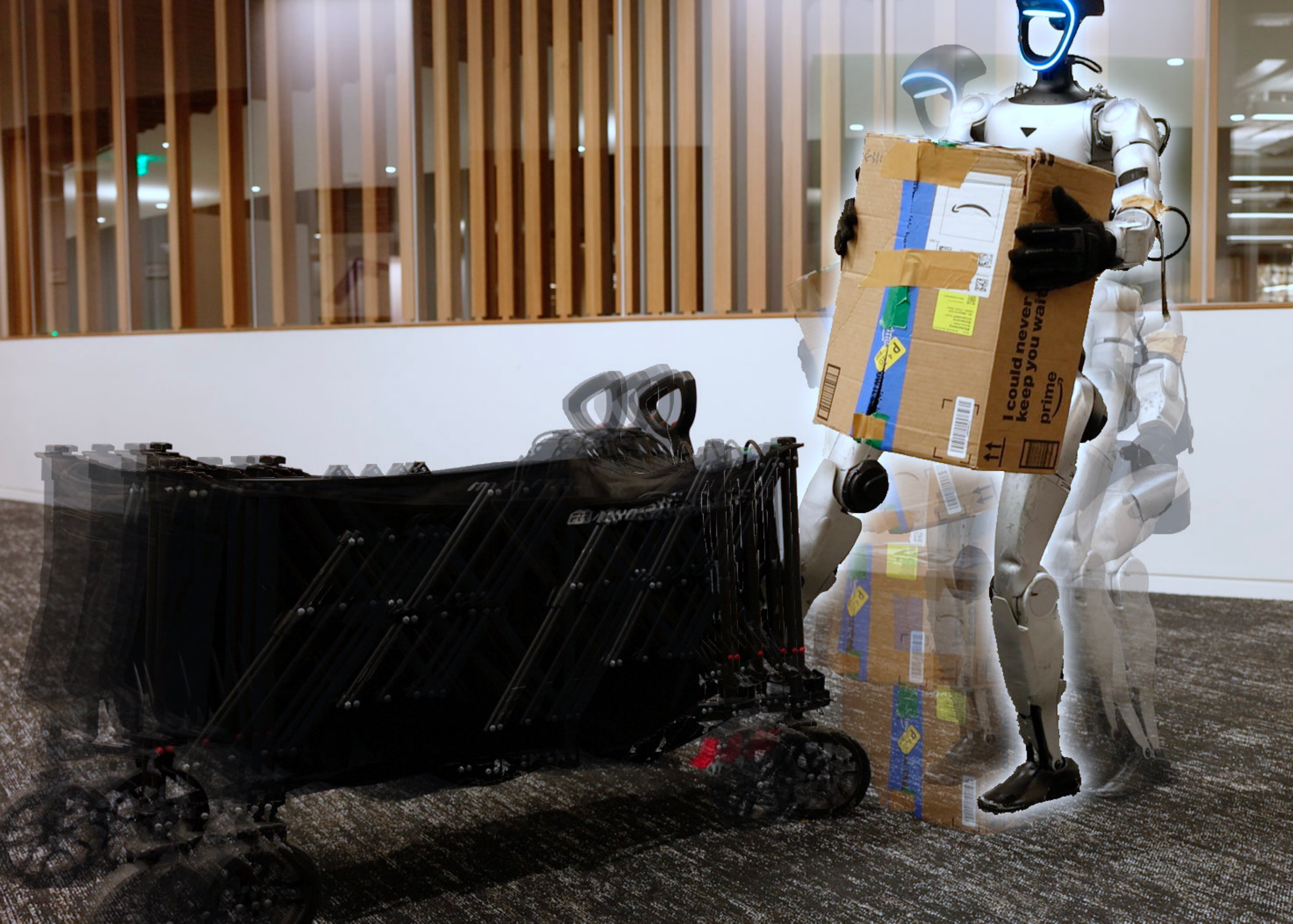}
\caption{Ctrl-F on hardware, running the raise-walk-kick sequence
with a $1$\,kg box in the hands and an $8$\,kg cart to kick. Raising the box is
already a force task: the hands slide off a box they do not press into, and a
tracker without a force command does not lift it at all.
}
\label{fig:teaser}
\end{figure}
\section{Introduction}

Many everyday whole-body tasks require sustained physical interaction with the environment, such as carrying a heavy box, dragging a crate, or pushing against a resistant door. These tasks demand coordinated whole-body motion and continuous force regulation as the environment responds. A heavier box requires a stronger grip, while a stuck crate demands greater pulling force.

Humanoid robots have made remarkable progress in reproducing human motion.
Policies have evolved from tracking individual clips~\cite{peng2018deepmimic} to general trackers capable of reproducing hours of motion capture with a single model~\cite{liao2025beyondmimic,luo2025sonic,gmt2025,qi2026humanoid}, leveraging the scalability of motion data. However, reference motions specify only where the body should move, without accounting for physical interaction. Even a strong motion tracker can fail when manipulating objects: in our experiments, a force-blind tracker fails to lift boxes weighing $1.5$–$6$\,kg, as the boxes slide off the robot’s palms, resulting in zero successful episodes. \emph{The motion is reproduced, but the intended interaction is not.}

Recent work has incorporated object interaction into the tracking objective through representations such as object motion, contact labels, and distance fields~\cite{weng2025hdmi,scenebot2026,lin2026lessmimic}. While these abstractions enable scalable learning, they primarily describe when and where contact occurs, rather than how much force should be exerted. Consequently, the policy cannot be explicitly commanded to apply greater force or adapt to heavier loads under the same reference motion. Force provides a natural and unified representation of physical interaction, directly capturing what the object experiences. Unlike discrete contact labels, force encodes both magnitude and direction, making it a continuous and expressive command signal for controlling physical interactions.

Force has been incorporated into humanoid control in several ways, though not as a general interaction command: as an external
disturbance resisted through force curriculum training~\cite{zhang2025falcon}, as end-effector stiffness modulated through hindsight goal perturbation~\cite{chen2025chip}, as a commandable position-force target at a single interaction point~\cite{zhi2025unified}.
While each addresses aspects of force control, none of them combines general motion imitation with explicit force commands \emph{and} online force adaptation based on object feedback during interaction.

To bridge this gap, we introduce \textbf{Ctrl-F}, a hierarchical humanoid loco-manipulation framework that explicitly commands the forces exerted on the environment. At the low level, a general motion–force controller jointly tracks a reference motion \emph{and} executes a 3-D force command for each designated contact body. Trained without objects through \emph{inverse-force training}, the controller preserves the scalability of general motion tracking without requiring object models or paired interaction data, while exposing force as an independently controllable input. Building on this frozen controller, a high-level force planner generates force commands from object states and proprioceptive observations. The planner is initialized through on-policy learning from an analytical force model and further refined via reinforcement learning, enabling it to adapt to contact dynamics beyond those captured by the analytical model.


Our contributions are:
\begin{list}{\arabic{enumi})}{%
  \usecounter{enumi}%
  \setlength{\leftmargin}{1.35em}%
  \setlength{\labelwidth}{1.05em}%
  \setlength{\labelsep}{0.3em}%
  \setlength{\itemindent}{0pt}%
  \setlength{\listparindent}{0pt}%
  \setlength{\topsep}{2pt}%
  \setlength{\itemsep}{2pt}%
  \setlength{\parsep}{0pt}%
}
\item We define a general force interface for humanoid interaction and train it without objects. Interaction is specified as a 3-D force at each designated contact body. The motion–force tracker is trained via inverse-force training without paired object assets or interaction data, enabling generalization across objects and interaction patterns without task-specific representations.
\item We train a high-level force planner to generate force commands from object and robot states. At deployment, the policy requires neither force measurements nor object mass, yet continuously adapts its commands to changes in object state and load.
\item 
We demonstrate sim-to-real deployment of our framework on a Unitree~G1 humanoid robot across multiple physical interaction tasks and loading scenarios, including online adaptation to load changes during execution.
\end{list}

\section{Related Work}

\textbf{Humanoid motion tracking.} DeepMimic established pose- and
velocity-imitation rewards with early termination and reference-state
initialization~\cite{peng2018deepmimic}. AMASS-scale data~\cite{mahmood2019amass}
enabled general trackers, including BeyondMimic~\cite{liao2025beyondmimic},
SONIC~\cite{luo2025sonic}, GMT~\cite{gmt2025} and
Humanoid-GPT~\cite{qi2026humanoid}. Such trackers also serve as real-time
teleoperation backbones, as in H2O~\cite{he2024h2o} and SONIC. These policies reproduce a kinematic
trajectory and are scored on their fidelity to it.
However, these policies primarily learn free-space motion, as neither the training objective nor the evaluation metric accounts for interaction with the environment. A tracker may perfectly reproduce a reference motion without exerting the forces needed to manipulate objects. Building on this foundation, we extend general motion tracking with explicit interaction capabilities rather than replacing it.

\textbf{Humanoid object interaction.} A second body of work takes the object
into account. HDMI co-tracks humanoid and object trajectories extracted from
video~\cite{weng2025hdmi}; SceneBot conditions a single general tracking policy
on per-link contact labels, with scene assets reconstructed from human
motion~\cite{scenebot2026}; LessMimic uses unified distance-field
representations for long-horizon, reference-free
interaction~\cite{lin2026lessmimic}.
\wnote{Optionally one clause on interaction-preserving
retargeting/frameworks, OmniRetarget and Holosoma, cited as infrastructure
rather than as force methods.}
These methods represent interaction through geometric quantities, such as object trajectories, contact labels, and surface distances. These representations can be extracted from human motion data at scale, enabling generalization across scenes. However, they capture where and when interaction occurs, rather than the force required. Without explicit force control, the policy cannot be commanded to push harder or adapt to heavier loads under the same reference motion.

\textbf{Force control, classical and learned.} Impedance
control~\cite{hogan1985impedance} and hybrid position/force
control~\cite{raibert1981hybrid} supply the mechanical vocabulary for
regulating contact, but not a recipe for preserving a high-dimensional motion
capture reference on a floating-base humanoid whose contacts change during the
motion. Among learned approaches, force has entered humanoid and legged
control in two main roles. One treats it as a disturbance: policies are
trained under external forces or randomized loads so that tracking stays
accurate when the robot is pushed or pulled~\cite{zhang2025falcon,dong2025hafo,sun2025ulc}.
FALCON, for example, trains dual lower/upper-body agents under a
torque-limit-aware external-force curriculum, making end-effector tracking
robust to applied wrenches~\cite{zhang2025falcon}. The other shapes the
response to contact through compliance, with a virtual spring or impedance
setting how the body yields~\cite{chen2025chip,xu2025facet,lu2025gentlehumanoid};
CHIP, for example, modulates end-effector stiffness on a generalist tracker
through hindsight goal perturbation~\cite{chen2025chip}. In the former role
force is a disturbance to reject rather than an objective to follow; in the
latter, force is regulated indirectly through stiffness and reference offsets, without closed-loop adaptation to object feedback.
UniFP unifies position and force control at a single interaction point
without force sensors, with a high level trained by imitation on
teleoperated, force-annotated data~\cite{zhi2025unified}. 
While force is explicitly commandable, the humanoid controller operates through base velocity and force commands without a whole-body motion reference, limiting the diversity of motions it can reproduce. More broadly, existing approaches either lack integration with general motion tracking or rely on predefined force commands without adapting to object feedback. Ctrl-F bridges this gap by combining general motion tracking with online force adaptation.

\begin{figure}[!tp]
\centering
\includegraphics[width=0.49\linewidth,trim=0 117 0 0,clip]{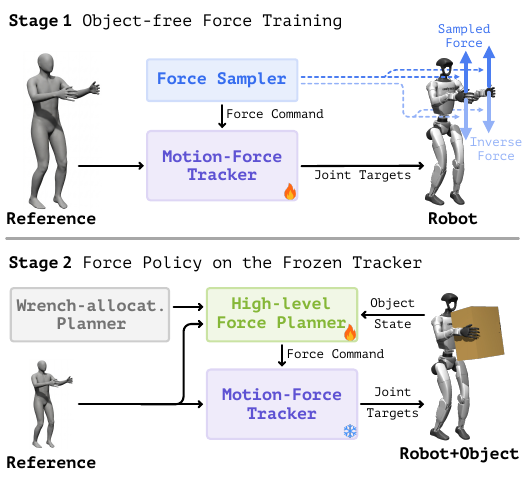}\hfill
\vrule\hfill
\includegraphics[width=0.49\linewidth,trim=0 0 0 116,clip]{figs/fig_method.pdf}
\caption{Ctrl-F. \emph{Stage 1} trains the motion--force tracker without objects: a force command is randomly sampled and its corresponding inverse force is applied to the
robot, teaching the tracker to follow its reference while exerting the commanded load. \emph{Stage 2} freezes that tracker and learns the force command itself
from object and robot state, supervised by the analytical force model and refined by
reinforcement learning.}
\label{fig:arch}
\end{figure}

\section{Method}

We propose a hierarchical framework (Fig.~\ref{fig:arch}) comprising a low-level motion-force hybrid tracker and a high-level force planner. The low-level tracker takes a robot reference motion, a
3-D force command per designated contact body, and proprioception as input, and outputs full-body joint-position targets. It has no notion of an object at all: the force command
is simply another input to track, allowing it to be trained without an object.
The high-level force planner takes as input the reference motions of both the robot and the object, the desired contact point on the object, the
object's measured state, and proprioception, and outputs the desired force command for the
low-level tracker. Proprioception and object state close the loop: by
watching how the object and robot body actually move against where the reference says it
should be, the high-level force planner decides how hard to ask the low-level tracker to push, grip,
or pull.

The force command is identical across tasks: one 3-D force
vector per designated contact body, the hands and a foot in our tasks, and, in
principle any body. The force command interface remains fixed across tasks,
allowing a single high-level force planner to span different interaction mechanics. The two levels
are trained sequentially: the low-level tracker is  trained without objects and then frozen, allowing the high-level planner to learn object interactions without relearning whole-body balance.

\subsection{Low-level motion--force tracker}
\label{sec:ll}

The low-level policy is a motion tracker with one addition: a
contact-force command input. Its objective is to maintain whole-body tracking while executing the force command.
\begin{figure}[!tp]
\centering
\includegraphics[width=0.245\linewidth,trim=0 109 126.5 0,clip]{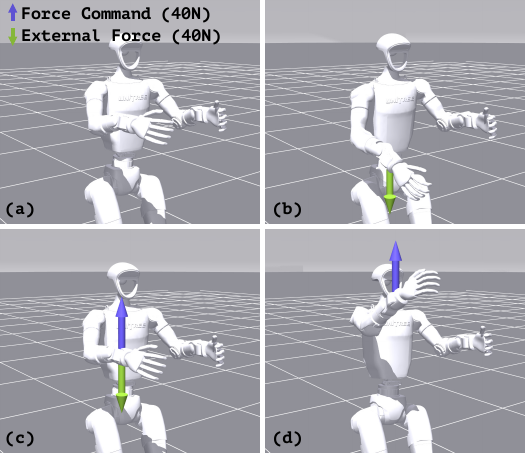}\hfill
\includegraphics[width=0.245\linewidth,trim=126.5 109 0 0,clip]{figs/force_command.pdf}\hfill
\includegraphics[width=0.245\linewidth,trim=0 0 126.5 109.5,clip]{figs/force_command.pdf}\hfill
\includegraphics[width=0.245\linewidth,trim=126.5 0 0 109.5,clip]{figs/force_command.pdf}
\caption{Response of the low-level tracker to an external wrench and a force command, during a static hold with a fixed reference posture.}
\label{fig:ll}
\end{figure}

\textbf{Inverse-force training.} 
The main challenge during low-level policy training is that no object is present: a
commanded force has nothing to act on. \emph{Inverse-force training}
handles this issue by turning force execution into a tracking problem. Each reference clip carries a per-body load mask marking when
each contact body is free and can exert force. Whenever that mask marks a body as free, we sample a force command and apply the \emph{opposite} wrench to that body while retaining the standard motion-tracking objective. To track the reference motion under this external wrench, the policy must produce an equal and opposite force, precisely the commanded
force. In this way, the policy learns to map a force command to the corresponding force-producing behavior without requiring an object during training. Because the learned response depends only on proprioception rather than object-specific observations or contact modeling, it transfers naturally from compensating for the injected wrench to exerting force through physical object contact.

Fig.~\ref{fig:ll} illustrates the learned force response on a static reference posture. Every panel in the figure is the trained motion-force tracking policy rollout in close loop, showing exactly the controller's behavior after training. Without an external wrench or force command, the policy holds the reference posture (a). An external wrench alone drives the arm away from the reference (b), while adding the matched force command causes the policy to counteract the wrench and recover the reference posture (c). With the command alone and no opposing wrench as in (d), the policy responds to the force command by producing that force against the world, and with nothing external to resist it, such action drives the robot body away from the reference motion and towards the force command direction, by approximately the same amount as in the opposite direction under the wrench alone. Together, these cases demonstrate the force response learned through inverse-force training.

\textbf{Load mask.} As mentioned previously, we use a per-body load mask to determine where and when force commands may be sampled. Importantly, the mask encodes physical feasibility rather than task
semantics: it does not identify when the reference motion is interacting with
an object, but only whether a body is in a configuration where it can
meaningfully exert an external force. For example, a supporting foot is masked
out because it cannot independently exert the commanded interaction force,
whereas the hands can remain available for force commands throughout most
reference motions. The mask can be derived directly from motion data through simple processing, preserving the data scalability of general motion tracking without requiring object interaction annotations.

\textbf{Curriculum.} We use a curriculum for learning motion-force tracking during the low-level training. Each contact body has a fixed range its force command is drawn from. During training, we first fix the force command to zero, hence no force is asked of the policy and it learns the basic motion-tracking ability. From a given training step onward, we gradually ramp the force command sampling scale to one, so that the policy meets small commanded forces first and the full per-body range last.

\subsection{High-level force planner}
\label{sec:hl}

In the second stage, the real object is put into the scene, and the high-level force planner is trained to give force command to the frozen low-level motion--force controller. That force is not observable from the reference motion: the same clip carried out on a light box and a heavy one needs different forces. However, learning it from reward alone is slow and unreliable, because a force
command is only scored through its effect on the object, and most commands
break contact before they produce a stable one. We therefore use an analytical force model as a teacher, and train the high-level force planner with an imitation loss on top of PPO. 

\textbf{Analytical force model supplies the force targets.}
\label{sec:planner}
The high-level force planner needs a source of physically sensible force targets. We
derive it from the object alone. We therefore use an analytical force model to give the coarse force targets. The analytical force model first computes the spatial
wrench $W^\star$ the object needs by computed-wrench control on the object:
proportional--derivative feedback on object pose and velocity error,
mapped through the object's Newton--Euler dynamics, plus feedforward
gravity compensation for airborne objects and Coulomb-friction
compensation for objects on the ground. It then allocates that wrench to
the available contact-force channels:
\begin{equation}
  \text{find } f \quad \text{such that} \quad G f \approx W^\star, \quad
  f \in \mathcal{F},
  \label{eq:alloc}
\end{equation}
here $W^\star\in\mathbb{R}^6$ is the desired object wrench just computed,
$f$ stacks the contact forces the robot may apply at its designated contact
bodies, $G$ maps those forces to the wrench they produce on the object, the
grasp matrix of grasp analysis~\cite{murray1994grasping,prattichizzo2016grasping}, and
$\mathcal{F}$ is the set of forces that channel can physically carry. A task enters this problem only as a triple $(G,\mathcal{F},\text{load})$ and
differs in nothing else. Every allocation is solved in closed form and batched across environments, allowing the
analytical force model to be queried at every step of every environment during
training.




The analytical force model reads privileged object state and true mass, hence it is not a
controller we could deploy.
Also, it does not need to be accurate. It is coarse guidance and not a target: it only has to put the learner in the right region of the command space,
reinforcement learning is free to leave it wherever the model is wrong, and
that is why a small set of interaction patterns is enough.

\textbf{Annealed imitation on top of PPO.}
The high-level force planner is trained against two signals at once. 
The analytical force model is queried at the states the
current policy visits, in the style of DAgger~\cite{ross2011dagger}, keeping
supervision on-policy. Its weight in the loss decays over training,
following DAPG~\cite{rajeswaran2018learning}, allowing the policy to depart
from the model's command wherever departure pays. 
The reinforcement objective is scored on the object rather than on the robot:
reward follows the object's reference and success is defined on the object,
not on how closely the robot matches the clip; the frozen tracker already
handles that. During training, the object mass is sampled within a range at the beginning of each episode and also changes within the episode.


\begin{table}[!b]
\centering
\caption{Objects. Mass denotes the range sampled during evaluation; friction denotes the coefficient at the task-relevant contact surface: the palms for the raised box and the floor for all other tasks.}
\label{tab:objects}
\footnotesize
\begin{tabular}{lccccc}
\toprule
 & Raised box & Tethered box & Kicked box & Pushed box & Dragged crate \\
\midrule
Mass (kg) & $1.8$--$6.0$ & $3.6$--$7.0$ & $3.0$--$6.0$ & $4.6$--$13.0$ & $7.0$--$8.8$ \\
Friction & $0.60$ & --- & $0.25$ & $0.38$ & $0.57$ \\
\bottomrule
\end{tabular}
\end{table}

\section{Experiments}
\label{sec:exp}

Our experiments answer four questions in simulation and one on hardware.
\textbf{Q1:} does the low-level motion--force controller realize an explicit force interface, and
at what cost to tracking? \textbf{Q2:} does the hierarchy complete
forceful tasks, and how does it scale with object mass relative to
force-blind controllers? \textbf{Q3:} does the commanded force adapt
online when the load changes? \textbf{Q4:} is model-supervised BC+RL
the right training recipe? \textbf{Q5:} does the full hierarchy transfer
to hardware?

\subsection{Experimental setup}

\begin{wrapfigure}{R}{0.50\textwidth}
\vspace{-6pt}
\centering
\includegraphics[width=\linewidth]{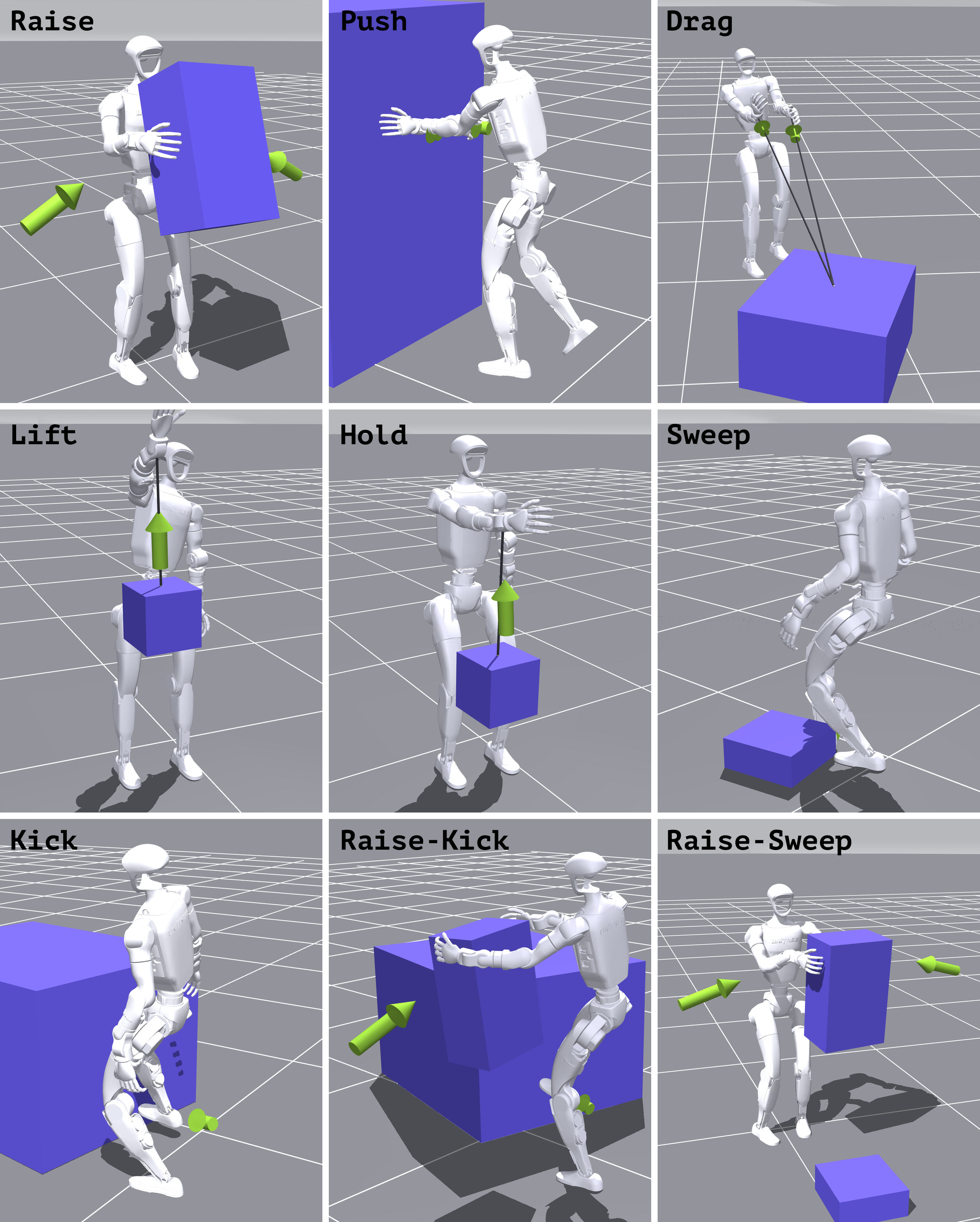}
\caption{The nine simulated tasks, in the reading order of the three families:
hand tasks raise, push, drag, lift and hold; foot tasks sweep and kick;
composite tasks carry a box while kicking or sweeping. Green arrows indicate the commanded force at each designated contact body, as specified by the high-level force planner and realized by the low-level controller.}
\label{fig:tasks}
\end{wrapfigure}

\textbf{Task families.} Our tasks fall into three families by which bodies
carry the load (Fig.~\ref{fig:tasks}). \emph{Hand tasks} raise, push, drag,
lift and hold a box with the hands. \emph{Foot tasks} kick and sweep with one
foot planted. \emph{Composite tasks} carry a box in the hands while kicking or
sweeping with a foot, loading wrists and an ankle at once. Each family has its own motion
corpus and its own controller pair, one low-level and one high-level shared by
every task in the family; a result within a family is therefore one policy
across several tasks rather than one policy per task.

\textbf{Objects.} Tab.~\ref{tab:objects} lists the objects the tasks act on.
Mass is randomized per episode within a band, and the sweep of
Sec.~\ref{sec:exp}-C reports it as a multiple of each band's midpoint, so
$1.0\times$ means the same relative position in every band even though the
absolute weights differ by a factor of two across objects.

\textbf{Baseline.} We compare against SONIC~\cite{luo2025sonic}, a general
humanoid motion tracker released with public G1 weights, run inside our
simulator, mjlab~\cite{zakka2026mjlab}, on the same clips and scored by the
same metrics.
SONIC serves as a representative force-blind baseline because it embodies the free-space position: a
strong tracker with no force channel at all.
We further evaluate an ablated version of our policy with the force command removed, keeping all other components unchanged to isolate the contribution of explicit force conditioning. 

\textbf{Metrics.} We report motion tracking in two frames throughout. \emph{Global error}
compares body positions to the reference in the world frame, and therefore
accumulates drift; \emph{local error} re-anchors the reference to the
robot's own root before comparing, and therefore measures posture fidelity
independently of where the robot ended up. We report each for the
whole body and, separately, for the contact bodies that carry
force: the bodies the interface acts on are exactly where a force-blind
baseline should suffer most. For the interaction tasks we also report
\emph{completion}, the fraction of episodes that reach the end of the clip
without a termination, and \emph{object error}, the 3-D distance between the
object and its reference position.

\subsection{The force interface (Q1)}

A force channel is worth having only if it is \emph{free when unused} and
\emph{effective when used}. We test the two halves separately against SONIC.

\begin{wraptable}{r}{0.45\textwidth}
\vspace{-10pt}
\centering
\caption{Q1(a): Tracking with the force command zeroed and no external load,
averaged across all twelve clips of the three task families. Errors are in cm;
``forced'' denotes the bodies specified by each clip's load mask. Local errors are root-anchored and global errors are measured in the world frame. Each family uses its own tracker; results are therefore averaged across clips,
with the two policies paired within each clip.}
\label{tab:q1a}
\footnotesize
\begin{tabular}{lcc}
\toprule
& Ours & SONIC \\
\midrule
Forced-body error, local & \textbf{2.96} & 5.89 \\
Forced-body error, global & \textbf{5.39} & 9.40 \\
Whole-body error, local & \textbf{3.10} & 3.75 \\
Whole-body error, global & \textbf{4.98} & 9.00 \\
Completion (\%) & \textbf{100.0} & 80.2 \\
\bottomrule
\end{tabular}
\end{wraptable}

\textbf{(a) Free when unused.} We evaluate motion tracking without external loads by setting the low-level force command to zero (Tab.~\ref{tab:q1a}). Adding the force channel does not compromise tracking performance: whole-body posture errors remain comparable, our forced-body error is roughly half that of SONIC, and our tracker completes all episodes. SONIC’s lower completion rate primarily stems from clips with root translation, where the absence of root-position input causes early termination even without external loads. However, two factors favor our tracker: it was trained on the evaluated clips, whereas SONIC encounters them for the first time, and SONIC uses its original G1 configuration rather than ours. These differences may exaggerate the performance gap. Thus, we interpret Tab.~\ref{tab:q1a} as evidence that force conditioning preserves motion tracking performance, rather than demonstrating superiority over SONIC.

\begin{figure}[!tp]
\begin{minipage}[b]{0.505\textwidth}
\centering
\includegraphics[width=\linewidth]{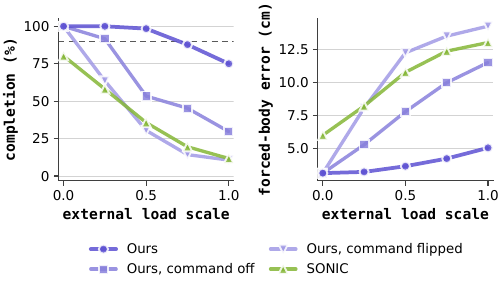}
\phantomsubcaption\label{fig:q1}
{\small (a)}
\end{minipage}\hfill
\begin{minipage}[b]{0.45\textwidth}
\centering
\includegraphics[width=\linewidth]{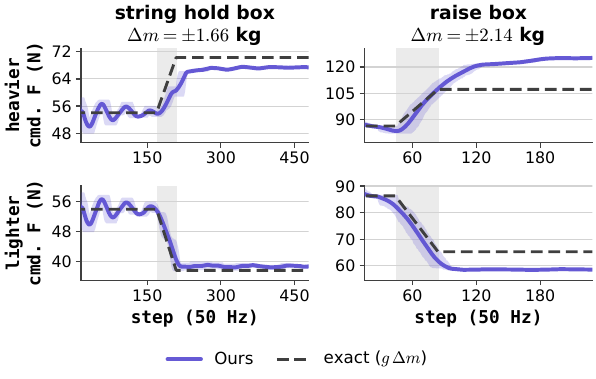}
\phantomsubcaption\label{fig:q3}
{\small (b)}
\end{minipage}
\caption{\textbf{(a)} Q1(b): completion and forced-body error as the replayed wrench is
scaled, averaged over all twelve clips. One trace per clip is replayed
identically to the four settings and scaled together with the command each is
given, so the settings differ only in the command they receive. The three command
variants are one policy and coincide at zero load. The scale is the fraction
of the per-body command range the wrench uses, so $1.0$ means the same thing
on a wrist ($60$\,N) and an ankle ($30$\,N) and nothing plotted is out of
distribution. Dashed line is the $90\%$ completion level. Errors are over all
active steps, a convention that flatters a setting that terminates early. \textbf{(b)} Commanded force through an in-episode mass change on the tethered
hold and the raised box, summed over force bodies. The solid line is a moving
average and the shaded band is the spread it removed, the minimum and maximum
within the same window; the grey column is the mass ramp; the dashed line
adds $g\,\Delta m$, the change in the object's weight, to the force before the
change. A tethered box is an undamped
pendulum and the commanded force oscillates with its swing, so the trend
rather than the raw trace carries the reading.}
\end{figure}

\textbf{(b) Effective when used.} We evaluate whether the low-level controller learns to exert the commanded force by measuring its effect on posture stability.
An external wrench is
applied to the robot's loaded bodies, as if it were carrying something, and the
force command is set to the force that counteracts it (Fig.~\ref{fig:ll}). A controller that executes the command should cancel the wrench and maintain the reference posture, whereas one without effective force control can only react to the disturbance, leading to drift and eventual failure. Thus, posture stability under load serves as an indirect measure of force tracking. This setup mirrors the low-level training procedure 
(Sec.~\ref{sec:ll}). 
However, a controller could also maintain posture by simply resisting disturbances without following the force command. To distinguish force conditioning from disturbance rejection, we compare policies under identical external wrenches while varying only the force command.
For each clip, the wrench is sampled once and replayed across all settings, with its magnitude scaled alongside the command. We evaluate four settings: our full policy, the same policy with the force command removed, the same policy with the command sign reversed, and SONIC, which has no force input (Fig.~\ref{fig:q1}).

The four settings show clear and consistent performance differences. Our policy degrades gradually, maintaining $75\%$ completion at full load, compared to $30\%$ or less for the other three. Its forced-body error remains within a few centimeters, while the others increase roughly threefold. The sign-inverted setting consistently performs worst: although the command magnitude is correct, its reversed direction causes the policy to reinforce rather than counteract the load, performing even worse than receiving no command. This confirms that the policy responds to both the magnitude and direction of the commanded force, rather than ignoring it. Walking clips remain the most challenging across all settings, with SONIC failing even without external loads.

\subsection{Task performance under increasing load (Q2)}

\begin{figure}[!tp]
\centering
\includegraphics[width=1.0\linewidth]{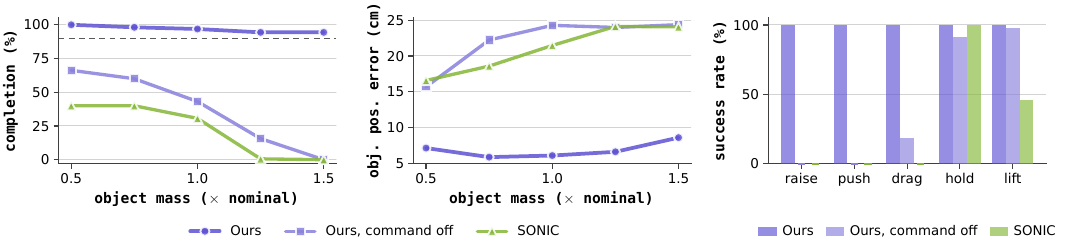}
\caption{\emph{Left:} Task performance as object mass increases, averaged over the five
hand tasks. Nominal is each object's trained-band midpoint, so
$1.0\times$ is the same relative position in every band.
Completion is over all active steps; a setting that terminates early is read together
with its survival. The dashed line is $90\%$ completion. \emph{Right:} success at nominal mass, per task, for the hand
policy (push: surviving $4$\,s); the tethered tasks have mirrored left and right clips and we draw the
right one of each.}
\label{fig:q2}
\end{figure}

We pin object mass to a multiple of each object's nominal value and measure
how task performance degrades as the mass grows, with and without the force
command (Fig.~\ref{fig:q2}), averaged over the five hand tasks.

With the force command, completion stays above $90\%$ across the swept range.
Without it, both force-blind policies hold for a while and then collapse just
past nominal mass: at nominal they complete $31\%$ (SONIC) and $43\%$ (ours
without the command), and by $1.5\times$ both are at zero. Neither reaches the
$90\%$ line anywhere in the range, not even at half nominal mass. Object error
separates in the same direction.

The two force-blind baselines share no weights: one is an external tracker, while the other is our policy with force commands removed. Their common limitation is the absence of force conditioning, rather than a shared training deficiency. Explicit force commands extend the operating range but do not eliminate the load limit; our policy also fails under sufficiently large loads.

The foot tasks show similar gains in load tolerance. Taking the mass at which completion
first falls below $90\%$, the kick holds out to $2.05\times$ nominal against
$1.51\times$ with the command removed, and the sweep stays above the line
beyond $3\times$, where the command-removed policy crosses it at $2.11\times$.

\subsection{Online force adaptivity (Q3)}

While the previous section varies object mass across episodes, here we change it within an episode to evaluate online force adaptation. We quantify adaptation gain as the change in commanded force per kilogram after the object settles. The policy has access to neither object mass nor force measurements.

The commanded force follows the load with the correct sign in every
environment of every task, in both directions (Fig.~\ref{fig:q3}). On the
tethered clips the commanded gain is $4.7$--$9.2$\,N/kg, below the $9.81$ a
suspended mass requires; the force actually delivered to the object closes the
gap, measuring $9.4$--$10.6$\,N/kg, since a rope carries the weight it holds.
On the pushed and dragged boxes the gain is $2$--$4$\,N/kg, the scale $\mu g$
sets for a sliding load. On the raised box the gain exceeds $9.81$, at
$18.2$\,N/kg for added and $12.9$\,N/kg for removed mass, because it is
summed over both palms, and $85$--$90\%$ of it is added inward pressing force:
the policy grips harder rather than lifting harder, which is what friction
requires.

This response does not depend on mass changing during training. A policy
trained with mass fixed within each episode adapts just as much. Both
policies observe the object's state and adjust the command when the object
starts to drift from its reference, which is enough to follow the load.

\subsection{Why both imitation and reinforcement (Q4)}

\begin{wraptable}{R}{0.47\textwidth}
\vspace{-10pt}
\centering
\caption{Training-recipe comparison over the five hand tasks. 3-D object
position error in cm, completion in \% in parentheses. ``no cmd.'' is our
policy with the force command removed; ``BC only'' is the same architecture
trained by behavior cloning alone. Where a task has mirrored left and right
clips we report the side our own policy does worse on.}
\label{tab:q4a}
\footnotesize
\begin{tabular}{lccc}
\toprule
Task & no cmd. & BC only & Ours \\
\midrule
Raise box       & 19.45 (0)    & 26.03 (6)    & \textbf{4.59} (100) \\
Tethered lift    & 25.25 (48)   & 23.65 (0)    & \textbf{6.26} (100) \\
Tethered hold    & 26.40 (75)   & 21.05 (84)   & \textbf{6.28} (100) \\
Push box        & 29.45 (0)    & 24.80 (0)    & \textbf{4.91} (84) \\
Drag crate      & 32.05 (19)   & \textbf{5.33} (100) & 6.95 (100) \\
\bottomrule
\end{tabular}
\end{wraptable}

\textbf{(a) Why reinforcement is needed.} Our high-level force planner is trained with PPO and an annealed imitation loss guided by an analytical force model (Sec.~\ref{sec:hl}). While the analytical model uses the object’s true mass and inertia, these quantities are unavailable to the deployed policy. We investigate whether imitation alone can recover effective force control from the available observations. To this end, we compare behavior cloning against our full training approach and a no-command variant with the force command removed (Tab.~\ref{tab:q4a}).

Imitation alone is insufficient. On four of the five tasks the behavior-cloned policy performs only marginally better than the no-command policy, despite closely matching the teacher’s commands on visited states. 
However, matching force commands does not guarantee effective force transfer to the object. For example, in the box-lifting task, the cloned policy commands 37\,N, but only about 8\,N is transmitted to the box.
The imitation loss cannot capture this discrepancy, as it supervises commanded forces rather than their effects on the object. 
Dragging is the exception: the crate requires a nearly constant pulling force determined by its weight and floor friction, which the analytical teacher already provides. Consequently, behavior cloning matches our full policy, with both achieving $100\%$ completion.
By optimizing task outcomes through reinforcement learning rather than solely matching teacher commands, our full approach achieves similarly high completion rates on the remaining four tasks.

\begin{figure}[!t]
\centering
\includegraphics[width=\linewidth]{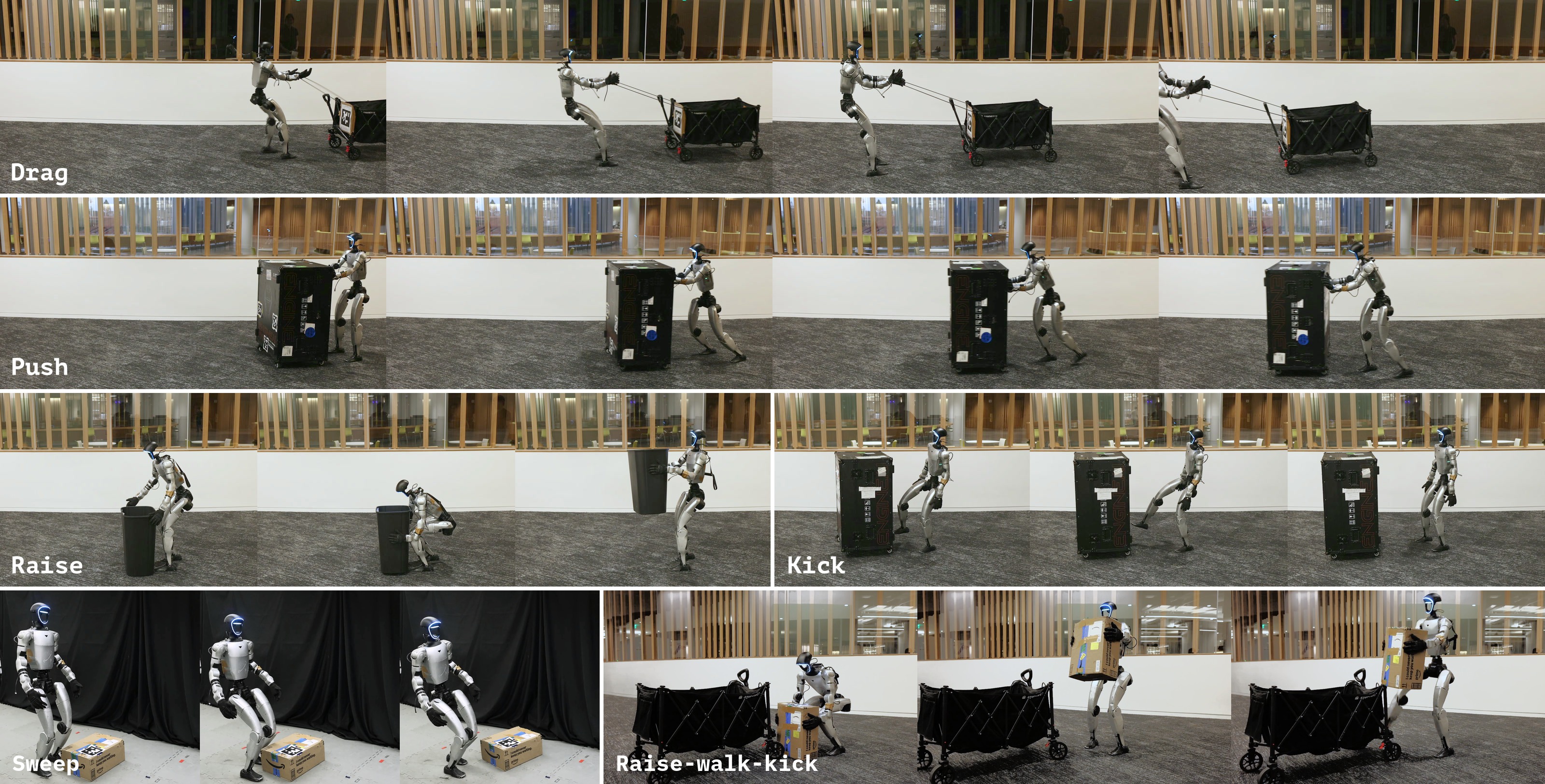}
\caption{Hardware deployment on a Unitree~G1 using onboard sensing and AprilTag-based object pose estimation, by task family. Hand tasks: dragging the cart, pushing the box and raising the bin. Foot tasks: kicking the box and sweeping a $1$\,kg box. Composite: raise-walk-kick, carrying a $1$\,kg box and kicking a cart loaded with $8$\,kg. The pushed and kicked boxes reach $20$\,kg, the heaviest load tested.}
\label{fig:q5}
\end{figure}

\textbf{(b) Why imitation is needed.} The reverse question does not admit a
controlled comparison: a policy trained without the analytical force model
fails differently from run to run. We report instead what recurred across our
attempts.
On clips that alternate between loaded and unloaded phases, reinforcement
learning alone collapses the distinction, commanding force throughout or not
at all, and which of the two it settles on depends on how the force samples
are drawn. On clips that are loaded from beginning to end it often reaches acceptable
task performance while commanding forces that bear little relation to what
the mechanics require. 
The analytical force model provides a structured prior rather than a reward signal: imitation establishes when and how much force to apply, while reinforcement learning refines these commands to account for the actual interaction dynamics.

\subsection{Transfer to hardware (Q5)}

\begin{wrapfigure}[12]{r}{0.46\textwidth}
\vspace{-6pt}
\centering
\includegraphics[width=\linewidth]{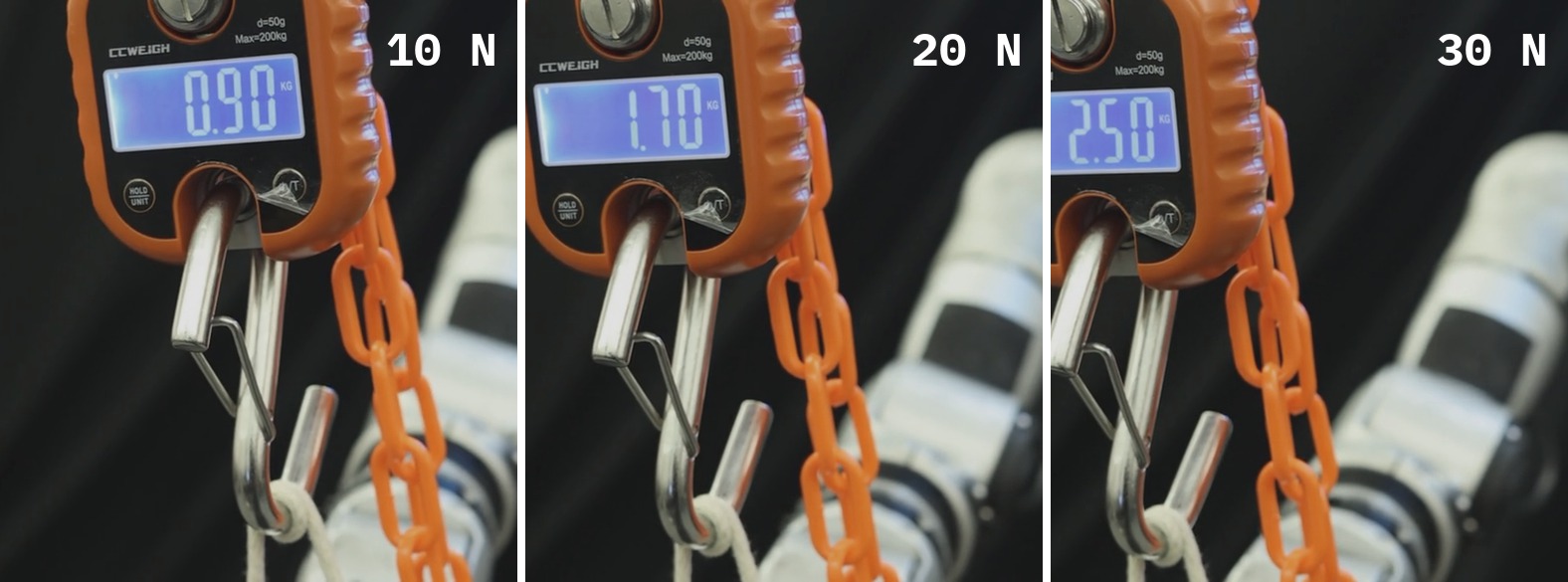}
\caption{Force realization on hardware. The hand pulls on a crane scale while
the low-level tracker is commanded $10$, $20$ and $30$\,N in one continuous
take; the scale reads $0.90$, $1.70$ and $2.50$\,kg ($8.8$, $16.7$ and
$24.5$\,N).}
\label{fig:gauge}
\end{wrapfigure}

We deploy the hierarchy on a Unitree~G1 robot with onboard sensing only and object
pose from AprilTag markers. Throughout, we compare against SONIC on the same
robot, objects and reference motions (Fig.~\ref{fig:q5}). The deployment
addresses three questions: whether the force command is realized as a physical
force, whether the hierarchy completes the task families of simulation, and
whether it adapts online when the load changes during a task, as it does in
simulation (Sec.~\ref{sec:exp}-D).

\textbf{Force realization.} We first test the low-level tracker alone. The
robot's hand pulls on a crane scale while the force command is stepped through
$10$, $20$ and $30$\,N in one continuous take (Fig.~\ref{fig:gauge}). The
scale reads $8.8$, $16.7$ and $24.5$\,N: the reading follows each step and
falls $12$--$18\%$ short of the command. Part of that gap is compliance at the
palm and in the cord, which the tracker does not model; part is the
measurement itself, a consumer crane scale pulled through a cord whose
direction is not exactly that of the commanded force, so the numbers indicate the scale of the
realized force rather than measure it precisely.

\begin{figure}[!t]
\centering
\includegraphics[width=\linewidth]{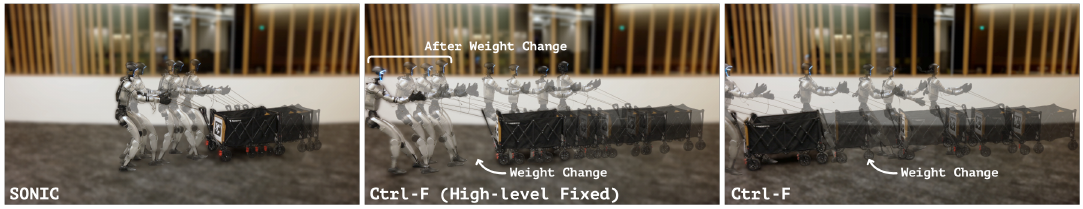}\\[2pt]
\includegraphics[width=0.993\linewidth]{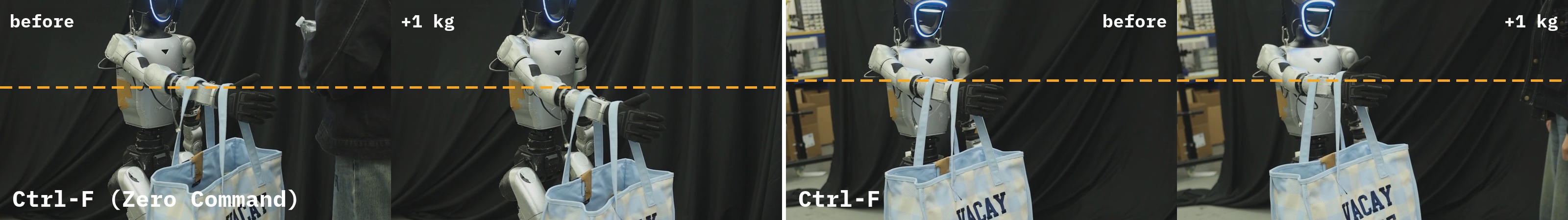}
\caption{Weight added mid-task on hardware. \emph{Top:} adapting to weight added mid-drag. In every setting the cart is loaded while
the robot is already dragging the cart, whose wheels stay locked throughout (arrow marks the
loading). Frames are composited at equal time intervals; the spacing between poses is
therefore speed. SONIC is given the easier test, an empty cart that takes
$8$\,kg, because it cannot start a loaded cart moving at all; even so, it stops
once the weight goes in. The two variants of our method start at $8$\,kg and take a further $6$\,kg. Under a
command frozen at what the high-level force planner itself issued for the initial load, the
robot keeps walking but slows, its poses bunching together after the change,
while the full hierarchy holds its pace. Those two share a tracker, an
interface and a starting force command, and differ only in whether that command is
allowed to change. \emph{Bottom:} An additional $1$\,kg is added to a $1$\,kg bag hanging from the forearm. The dashed line indicates the wrist height before loading. With zero force command, the wrist drops below this line, whereas the high-level force planner adapts its command to maintain the original height.}
\label{fig:ablation}
\end{figure}

\textbf{Hand tasks.} We deploy four: dragging a cart, pushing a box, raising a
bin, and holding a bag hung on the forearm. The cart's wheels are locked so the
robot drags it across carpet rather than rolling it, making its load sliding
friction. SONIC moves the empty cart and nothing heavier, stepping in place
under any added weight; ours drags it empty, with $8$\,kg and with $14$\,kg,
and fails at $20$\,kg. On the push box, SONIC does not move it either empty or
loaded with $20$\,kg, while ours pushes both at no obvious difference in
speed, the behavior a command that scales with the object should produce. On
the bin, SONIC raises the empty one because its hands catch the rim and fails
with $2$\,kg inside, its hands sliding past: nothing presses inward, and the
lip alone will not hold the load. Ours raises the $2$\,kg bin in $9$ of $10$
trials, commanding an inward force whose normal pressure is what produces
grip. With $1$ and with $2$\,kg in the bag, SONIC's hand sinks below its
reference under the load, while ours holds it at the reference height.

\textbf{Foot tasks.} We deploy the kick and the sweep. SONIC performs neither
once the object resists, being unstable on one leg: on the sweep it sways in place and never moves the $1$\,kg box. Ours kicks the box both empty
and with $20$\,kg in it, and sweeps the box clear (Fig.~\ref{fig:q5}).

\textbf{Composite task.} We deploy raise-walk-kick, which loads the hands and
an ankle within one motion: the robot carries a $1$\,kg box while kicking a
cart loaded with $8$\,kg, and performs the sequence successfully
(Fig.~\ref{fig:q5}).

\textbf{Mid-episode loading.} In the preceding tasks, the load remains constant throughout each episode. We further evaluate online adaptation through two tests that introduce additional weight during execution, assessing whether the hierarchical controller adjusts to the changing load. 
In the first test, we add weight to a cart during dragging (Fig.~\ref{fig:ablation}, top). The full hierarchy maintains its speed despite the increased load. As a baseline, we freeze the high-level force planner after two seconds, holding its command at the median of its recent outputs. This preserves the same tracker, force interface, and initially selected command, but prevents further adaptation. Consequently, the baseline slows down when additional weight is introduced, demonstrating that adapting to changing loads requires online force command updates beyond the low-level tracker’s capabilities.
In the second test, an additional $1$\,kg is added to the $1$\,kg bag on the forearm
(Fig.~\ref{fig:ablation}, bottom). With the high-level force planner, the hand maintains its reference height, whereas the same low-level controller with zero force command allows the hand to drop under the increased load. Together, these hardware tests validate the in-episode adaptation demonstrated in Sec.\ref{sec:exp}-D, showing that force commands adjust online to changing loads.

\section{Limitations}

The high-level force planner requires object-state estimates, obtained on hardware from AprilTag markers. This introduces deployment overhead and makes the system dependent on reliable marker detection.
Future work may explore using depth image as object observation.


Moreover, our policies are trained on a limited set of motions. Although a single policy supports diverse interaction mechanics, including unilateral pushing, frictional grasping, rope tension, and impact, it does not yet generalize to unseen motions. Unlike general motion trackers, the policy cannot follow a new motion clip without retraining. Scaling this interface to a large motion corpus is therefore a natural next step toward a general motion–force controller.

\section{Conclusion}

We presented a hierarchy framework for humanoid object interaction that explicitly controls the robot's interaction with the environment through force. Inverse-force training equips a general motion tracker with explicit force commands at arbitrary contact bodies without requiring objects during training. 
Above the frozen tracker, a high-level force planner specifies force commands from object and robot state  in a closed-loop manner. The planner is supervised by an analytical force model that unifies four
interaction mechanics and further refined through reinforcement learning. The analytical force model uses privileged states only during training; deployment requires only the tracker and learned force planner.

In simulation, the interface costs little tracking when idle, holds its
reference under an external load by exerting back the opposing force it is
commanded, and adapts online to mass changes it never observes. It holds
completion above $90\%$ out to $1.5\times$ nominal object mass, where both
force-blind policies have fallen to zero; at nominal mass those two stand at $43\%$
and $31\%$. Through an in-episode mass step the commanded force moves with the
load, with the correct sign in every environment. On hardware the robot produces forces close to the commanded ones on a crane
scale, and the same policies raised, dragged, pushed, held, swept and kicked loads the
force-blind baseline could not move, and absorbed weight added in the middle of
a drag and of a held bag. 
Future work will extend the framework to more diverse object classes and marker-free object representations.

\newpage
\balance

\end{document}